%% file: main.tex
\documentclass{article} 

\usepackage{aaai23-r2hcai}  

\usepackage{times}  
\usepackage{helvet}  
\usepackage{courier}  
\usepackage{url}  
\usepackage{hyperref}
\usepackage{graphicx} 
\usepackage{natbib}  
\usepackage[labelfont=bf,font=bf]{caption}
\usepackage{multirow}
\usepackage{amsmath}
\usepackage{amsfonts}
\usepackage{bm}
\usepackage{booktabs}
\usepackage[a4paper, total={7.0in, 9.30in}]{geometry}

\usepackage{fancyhdr}
\fancypagestyle{firstpagehf}
{
   \fancyhf{}
   \fancyfoot[C]{\thepage}
}

\title{Measuring the Effect of Causal Disentanglement on the Adversarial Robustness of Neural Network Models}
\author{
    Preben M. Ness \quad
    Dusica Marijan \quad
    Sunanda Bose
}
\affiliations {
    Simula Research Laboratory \\
    Oslo, Norway \\
    prebenmn@simula.no, dusica@simula.no, sunanda@simula.no
}

\begin{document}
\thispagestyle{firstpagehf}
\maketitle

\begin{abstract}
Causal Neural Network models have shown high levels of robustness to adversarial attacks as well as an increased capacity for generalisation tasks such as few-shot learning and rare-context classification compared to traditional Neural Networks. This robustness is argued to stem from the disentanglement of causal and confounder input signals. However, no quantitative study has yet measured the level of disentanglement achieved by these types of causal models or assessed how this relates to their adversarial robustness.

Existing causal disentanglement metrics are not applicable to deterministic models trained on real-world datasets. We, therefore, utilise metrics of content/style disentanglement from the field of Computer Vision to measure different aspects of the causal disentanglement for four state-of-the-art causal Neural Network models. By re-implementing these models with a common ResNet18 architecture we are able to fairly measure their adversarial robustness on three standard image classification benchmarking datasets under seven common white-box attacks. We find a strong association (r=0.820, p=0.001) between the degree to which models decorrelate causal and confounder signals and their adversarial robustness. Additionally, we find a moderate negative association between the pixel-level information content of the confounder signal and adversarial robustness (r=-0.597, p=0.040).
\end{abstract}

\section{Introduction}\label{sec:intro}
The latent internal data representations of a model are said to be disentangled when different signal components or dimensions model separate semantic concepts in the input. For a dataset of face images, this could mean separate signal components for e.g. \textit{gender}, \textit{age}, and \textit{presence of moustache}. It is a commonly held notion that such disentangled representations in Neural Network (NN) models are beneficial for the model's ability to adapt to new tasks or data distributions, decrease sample complexity, and increase the model's robustness to adversarial attacks \citep{disentangled_representations_good, disentangling_good_for_sample_complexity, adversarial_robustness_through_disentanglement}. However, the extensive investigation conducted in \citet{challenging_common_assumptions_disentangled_representations} challenges these broad general assumptions and highlights the importance of more specific studies quantifying the concrete benefits of disentangled representations for different tasks and desirable model attributes. 

\textit{Causal} disentanglement is a special type of disentangled representations where the aim is to separately represent input features which are \emph{causally} related to some output label and features which are merely \emph{spuriously correlated} with the label. For an image classification task, this could mean separately representing the subject of an image - e.g. \textit{a cat} - from the information about the background, lighting levels, or camera angle. The latter is often correlated with the image label - e.g. images of wild animals tend to have nature backgrounds - but this is not the \emph{cause} of the label, and hence this pattern might not generalise to unseen tasks or datasets. It is demonstrated in \citet{disentangled_representations_helpful_for_visual_reasoning} that disentangled representations reduce sample complexity for specific abstract visual reasoning tasks which were intentionally difficult to solve based purely on statistical co-occurrences of depicted objects. Furthermore, it is a commonly held belief that adversarial attacks exploit spurious or non-causal correlations learnt by a model \citep{kilbertus2018generalization}, it is therefore argued by e.g. \citet{toward_causal_representation_learning} that causal disentanglement should make models more robust against such attacks.

There is a class of Neural Network (NN) models which explicitly aim to achieve causal disentanglement using the mathematical framework of Causal Inference \citep{pearl2009causality}, throughout this paper we will refer to such models as \emph{Causal NNs}. These models have demonstrated good generalisation capabilities, and have been used successfully for long-tailed classification \citep{long-tailed_classification_causal}, to improve adversarial robustness \citep{vit_causal_intervention}, and to decrease sample complexity during training \citep{causal_disent_good_for_sample_complexity}. Although this performance is argued to stem from the models' ability to learn causally disentangled representations, there is a lack of studies investigating this claim. To the best of our knowledge, this is the first work to quantitatively test the association between causal disentanglement and adversarial robustness for NN models.

Since a good disentangled representation is taken to mean one where there is a correspondence between signal components and high-level semantic content in the input data, quantitative investigations have so far primarily been confined to synthetic datasets \citep{robustly_disentangled_causal_mechanisms}. This is because such a dataset allows one to both vary and measure the values of the true underlying data-generating factors. One can then vary a single factor - e.g. \textit{presence of moustache} - and confirm both qualitatively and quantitatively that only a subset of the representation's components varies while the rest are unchanged. This work is concerned with models operating on real-world datasets where the true values of the data-generating factors are of course unknown. Therefore, we propose a framework for measuring causal disentanglement using metrics based only on the information content and co-dependence of different representation components. This allows us for the first time to quantify the level of causal disentanglement achieved by state-of-the-art Causal NNs trained on real-world datasets, as well as measure the association between this disentanglement and each model's adversarial robustness \footnote{Code for all models and experiments available at \url{https://github.com/prebenness/causal_disentanglement_robustness}}.

\paragraph{Contributions}
\begin{itemize}
    \item We perform systematic benchmarks of four recent causal NN models across three standard datasets with a common ResNet18 backbone allowing for a fair comparison of the models' performance and robustness.
    \item We introduce a framework for quantifying causal disentanglement which does not depend on access to data-generating factors or stochastic model signals.
    \item We find that the degree to which the different models achieve separation of causal and confounder signals varies significantly, but is largely independent of dataset.
    \item We find a strong positive association between the decorrelation of causal and confounder signals and model robustness to adversarial attacks.
\end{itemize}

\section{Causal Neural Networks}\label{sec:causal_inference}
Throughout this paper, a Neural Network model is said to be \emph{causal} if it aims to explicitly separate the causally linked and spuriously correlated information contained in an input $\mathbf{x}$ with respect to some label $\mathbf{y}$. We denote the causal signal $\mathbf{c}$ and the spurious - or confounder - signal $\mathbf{s}$. Finally, any applied perturbation to the input data - e.g. an adversarial attack - is denoted $\mathbf{m}$ and the resulting perturbed input is denoted $\mathbf{\tilde{x}}$. Lowercase bold letters indicate vectors or tensors and upper case letters indicate random variables.

A key assumption in classical NNs is that training and test data samples are drawn from the same data distribution. This causes degradation in performance when there is a shift in the distribution of data between the train and test domains. The motivation behind Causal NNs is to learn the causal features and relationships which hold true across such shifts in the data distribution, hence improving the model's ability to generalise. As a result, this class of models has seen an increase in popularity over the past few years for use cases such as long-tailed classification \citep{long-tailed_classification_causal}, learning feature importance \citep{causal_feature_learning}, and defence against adversarial attacks \citep{causal_intervention_nlp}.

Subject to a successful disentanglement of the causal signal $\mathbf{c}$ and the confounder signal $\mathbf{s}$, the central mathematical operation in most Causal NNs is the \emph{back-door adjustment}. This is formalised in \citet{pearl2009causality} as the \emph{do-calculus} operation $P(Y|do(X)$. For a classifier predicting a label $Y$ from an image $X$ this becomes a marginalisation over the confounding variable given by

\begin{equation}
    P(Y|do(X)) = \sum_\mathbf{s} P(Y|X,S=\mathbf{s})P(S=\mathbf{s}),
    \label{eq:back_door_adjustment}
\end{equation}

\noindent where $\mathbf{s}$ is the confounding signal, e.g. style and background information in the image. We can now see that casual NNs aim to provide robust classifications by smoothing out any learnt spurious correlations between $S$ and $Y$. Although the application of Equation \ref{eq:back_door_adjustment} removes dependence on the confounder signal $\mathbf{s}$, any practical implementation is necessarily approximate. Firstly, the summation over all possible values of $\mathbf{s}$ is of course intractable, and in practice only finitely many terms can be used. Secondly, the isolation of $\mathbf{s}$ depends on the model achieving causal disentanglement to a sufficient extent.

\subsection{Causal Disentanglement}
No universally agreed-upon definition of \emph{disentangled representations} exists in the context of NNs \citep{definition_of_disentangled}. The term is generally taken to mean that semantically distinct components of an input are represented as separate components or dimensions of the model's internal representations. \emph{Causal} disentanglement has a narrower meaning in that the separate signal components represent the information in the input which is causally linked to the output and the information which is only spuriously correlated with the output for a given dataset. For real-world datasets where the true data-generating process is unknown and inaccessible, the definition of causal disentanglement must necessarily be qualitative. In this work we investigate image classification models, and we take the causal information to be the information defining the image subject as given by the label $\mathbf{y}$. We then take the spurious information to be the remaining information in the image, such as background, lighting, camera angle, and lens distortions. This is in line with the desired information content described in the works proposing our studied models.

It is proven by \citet{challenging_common_assumptions_disentangled_representations} that fully unsupervised learning of disentangled representations is impossible. Disentanglement must be enforced and encouraged by the choice of \emph{inductive biases}, e.g. the model architecture, choice of loss function and training regime, and sample weights and dataset splits. Causal NNs are of course subject to the same limitations, and the implementation and modelling choices made are crucial in achieving the desired causal disentanglement. We therefore here highlight three important design parameters for Causal NNs. In Section \ref{sec:the_models} we describe how the models we have investigated realise these parameters.

\subsubsection{Separation Mechanism}
In order to split the signal representation into the $C$ and $S$ components a dedicated separation mechanism is almost universally used in causal NN architectures. This can be as simple as a feedforward network with two outputs, but restrictions are often used to ensure that the two signal streams are in some sense complementary. Examples include using two orthogonal projection matrices \citep{CausalAdv}, an attention mechanism $a(\mathbf{x})$ and its complement $\mathbf{1} - a(\mathbf{x})$ \citep{CaaM}, and disjoint input masks based on measures of pixel classification importance \citep{DICE} \citep{CONTA_semantic_segmentation}.

\subsubsection{Intervention Mechanism}
As shown by \citet{pearl2009causality}, in order to identify causal signal components a so-called \emph{intervention} is necessary - in the case of a classifier the do-calculus operation $P(Y|do(X)$ as defined in Equation \ref{eq:back_door_adjustment}. In a physical approximation, this would correspond to e.g. collecting images of a target class under all possible lighting conditions, camera angles, etc, in order to evaluate the terms in the marginalisation sum. This is obviously practically impossible, and Causal NNs must therefore approximate the evaluation of Equation \ref{eq:back_door_adjustment}. We refer to the part of the model architecture that implements this approximation as the model's \emph{intervention mechanism}. Some models move the intervention mechanism to the model's latent space and use additive noise $\mathbf{n} \sim \mathcal{N}(\mathbf{0}, \mathbf{I})$ to approximate different confounder signal values as $\hat{\mathbf{s}} = \mathbf{s} + \mathbf{n}$ \citep{CausalAdv} \citep{causal_intervention_nlp}. Others such as \citet{CaaM} iteratively partition the training data during training with the aim of grouping input samples with similar confounder signal values into the same partition stratum.

\subsubsection{Auxiliary Loss}
While the purpose of a model's separation mechanism is to enforce the independence between the causal signal $C$ and the confounder signal $S$, it is also necessary to apply an inductive bias to enforce the desired information content of each signal stream with respect to the input. The $C$ signal is very often used as the basis for the model's primary task and can therefore be trained in the traditional way with a standard loss function. However, models differ in the choice of the auxiliary task associated with the confounder signal stream. Some employ $S$ in an adversarial way to select or create augmented training samples \citep{DICE} \citep{CaaM}, while others use $S$ directly for the primary task \citep{CausalAdv} in order to align the model's output distributions for clean and adversarially perturbed data.

\subsection{The Investigated Models}\label{sec:the_models}
In this paper, we study the following four models: the \emph{deep causal manipulation augmented model} (CAMA \citep{CAMA}), the \emph{causal attention module} (CaaM \citep{CaaM}), the \emph{causal-inspired adversarial distribution alignment method} (CausalAdv \citep{CausalAdv}), and the \emph{domain-attack invariant causal learning} model (DICE \citep{DICE}). Next, we give an overview of these models' architectures and design choices.

\subsubsection{CAMA}
Based on a Variational Auto-Encoder (VAE) architecture, CAMA aims to model the causal variables $M$ and $S$ through separate encoder networks. For clean training samples, the manipulation variable $M$ is set to a null value, and horizontally and vertically shifted images are used during training to model manipulated data. Similar to a standard VAE, the model aims to maximise the Evidence Lower Bound (ELBO) of the training data \citep{original_vae_paper}, which corresponds to $\sum_{\mathbf{x}, \mathbf{y}}ELBO(\mathbf{x},\mathbf{y}, \mathbf{m}=\mathbf{0})$ for clean data samples and $\sum_{\mathbf{x}, \mathbf{y}}ELBO(\mathbf{x},\mathbf{y})$ for manipulated data. 
 
\subsubsection{CaaM}
The original use-case of CaaM was to perform rare-context image classification on the datasets NICO \citep{nico_original_paper} and ImageNet-9 \citep{imagenet9_original_paper}. However, the model design utilises causal-confounder separation to the same end as the other models studied, namely to find distribution-invariant causal image features. The model uses a separation mechanism consisting of a CBAM \citep{woo2018cbam} attention mechanism $\mathbf{z} = \text{CBAM}(\mathbf{x})$ and its complement. The input $\mathbf{x}$ is separated into causal features $\mathbf{c}$ and confounder features $\mathbf{s}$ by the relations

\begin{align*}
    \mathbf{c} &= \text{Sigmoid}(\mathbf{z}) \odot \mathbf{x},\\
    \mathbf{s} &= \text{Sigmoid}(-\mathbf{z}) \odot \mathbf{x} = \bigl(\mathbf{1} - \text{Sigmoid}(\mathbf{x})\bigr) \odot \mathbf{x},
\end{align*}

\noindent where $\mathbf{z} \in \mathbb{R}^{w\times h \times c}$ and $\odot$ is the elementwise product. The confounder features $\mathbf{s}$ are then used to create a dataset partition $\tau$ of splits $t$ with similar confounder signal values which are used to approximate the backdoor adjustment of Equation \ref{eq:back_door_adjustment} as $P(Y|do(X)) \approx \sum_{t \in \tau} P(Y|X,t)P(t)$.

\subsubsection{CausalAdv}
The overall goal of CausalAdv is to align the modelled distributions of natural data $P(Y|X,s)$ and adversarial data $P(Y|\tilde{X},s)$. The input signal $\mathbf{x}$ is embedded to a latent space representation by a ResNet18 backbone to create $\mathbf{h} = \text{ResNet}(\mathbf{x})$. A trainable linear projection $\mathbf{W}_c$ is then used to extract the causal signal $\mathbf{c} = \mathbf{W}_c\mathbf{h}$. In order to separate out the confounder signal $\mathbf{s}$, a projection matrix $\mathbf{W}_s$ is constructed so that it is orthogonal to $\mathbf{W}_c$ in the sense that $\mathbf{W}_c\mathbf{h} \perp \mathbf{W}_s \mathbf{h}$ for all $\mathbf{h}$. As an approximation to the marginalisation over $\mathbf{s}$ in the backdoor-adjustment of Equation \ref{eq:back_door_adjustment}, random noise $\mathbf{n} \sim \mathcal{N}(\mathbf{0}, \sigma \mathbf{I})$ is added to produce $\hat{\mathbf{s}} = \mathbf{s} + \mathbf{n}$.

The distribution alignment is then approximated by a cross-entropy (CE) loss, with two classifiers $h$ and $g$ predicting sample labels from $\mathbf{c}$ and $\hat{\mathbf{s}}$ respectively. This loss is summed across both adversarial and natural samples as $\mathcal{L} = \alpha \text{CE}(h(\mathbf{c}), \mathbf{y}) + \beta \text{CE}(g(\hat{\mathbf{s}}), \mathbf{y}),$ where $\alpha$ and $\beta$ are positive real-valued scaling factors to adjust the relative weights of the different loss terms.

\subsubsection{DICE}
Similarly to CausalAdv, DICE also employs adversarial training to increase robustness. However, unlike the other models studied, DICE achieves the separation of causal and confounder signals through input masking. This mask is constructed by using the loss gradient $\delta \in \mathbb{R}^{w\times h \times c}$ of a reference classifier with respect to the pixels in the input image $\delta = \nabla_x\mathcal{L}(f_{ref}(\mathbf{x}), \mathbf{y})$. Pixels for which $\max_k\delta_{ijk}$ is above some threshold value are set to $0$ in order to produce a confounder sample $\mathbf{s}_x$. In order to approximate the marginalisation over all possible confounders, DICE utilises a finite replay buffer of generated confounder samples $\mathbb{S}$ and approximates backdoor-adjustment as

\begin{equation}
    P(Y|do(X)) \simeq \sum_{s \in \mathbb{S}} P(Y|X,s)P(s).
    \label{eq:dice_backdoor}
\end{equation}

\subsection{Adversarial Attacks}
Even state-of-the-art NN models are susceptible to performance degradation when the input is perturbed, often only very slightly so as to be virtually imperceptible to a human observer \citep{useful_not_robust_features}. Although the defence against such attacks is still an ongoing subject of research, a prevalent hypothesis in the field of Causal NNs is that adversarial attacks exploit learnt spurious correlations between $\mathbf{s}$ and $\mathbf{y}$ \citep{toward_causal_representation_learning}. NNs are extremely adept at capturing statistical relations but, unlike humans, lack an understanding of causal relations. As a result, carefully crafted changes to an input image targetting the confounder signal $\mathbf{s}$ can lead to misclassifications in a NN while being completely ineffective against humans. Since Causal NNs aim to correctly learn the causal relations between input and output data, it is argued that they can circumvent this adversarial attack vector. In order to measure the adversarial robustness of the investigated models we subject them to a range of common attacks, these are outlined in this section. 

All attacks are so-called \emph{white-box} attacks, where the attacker has full access to the weights $\bm{\theta}$ and loss gradients $\nabla\mathcal{L}(\bm{\theta}, \mathbf{x}, \mathbf{y})$ of the attacked model. White box attacks are therefore considered the most difficult attack types to defend against. The perturbations generated by the attacks are constrained to lie within a ball of a small radius $\epsilon$ around the clean sample, that is $||\mathbf{x} - \mathbf{\tilde{x}}||_p \leq \epsilon$, where $||\ .\ ||_p$ denotes the $l_p$ norm of a vector or tensor.

\subsubsection{Projected Gradient Descent}
Originally proposed in \citet{pgd}, Projected Gradient Descent (PGD) is an iterative perturbation scheme which at each iteration step $t$ applies a small perturbation $\delta_t$ to an input image $\mathbf{x}$ in the direction of the loss function gradient $\nabla_{\mathbf{x}}\mathcal{L}$. The new image $\mathbf{x}_t = \mathbf{x}_{t-1} + \delta_t$ is then clipped to a ball of radius $\epsilon$ under the chosen distance norm in order to ensure that the total allowed perturbation relative to the original input is not exceeded. The algorithm then iterates for a pre-specified number of steps or until a convergence criterion is met.

\subsubsection{CW}
Similar to PGD, the attack method CW proposed in \citet{cw} is an iterative optimisation-based scheme, but the objective in this context is to jointly maximise the discrepancy between the true and predicted label, and minimise the perturbation distance relative to the original image. This is achieved by optimising a surrogate compound loss function using e.g. gradient descent for a specified number of iteration steps.

\subsubsection{FGSM}
Both PGD and CW are effective attack methods used to test the robustness of state-of-the-art adversarial defence methods, but due to their iterative formulations, they are comparatively computationally expensive. In contrast, the Fast Gradient Sign Method (FGSM) \citep{fgsm} calculates a single perturbation proportional to the sign of the model's loss gradient as $\delta = \epsilon\ \text{Sign}(\nabla_{\mathbf{x}}\mathcal{L})$. Although not as effective as PGD and CW, FGSM is a popular attack algorithm due to its lower computational cost.

\subsection{Disentanglement Metrics}
Quantifying disentanglement in NNs is motivated by the heuristic idea that in a disentangled representation different signal components should correspond to different high-level semantic concepts in the data represented. Although a multitude of quantitative disentanglement metrics has been proposed \citep{review_disentanglement_metrics} \citep{disentangling_by_factorising}, the vast majority are restricted by at least one of the following two strong assumptions. 

Firstly, a large body of work on disentanglement quantification is concerned with models trained on synthetically generated datasets \citep{challenging_common_assumptions_disentangled_representations} \citep{disentangling_by_factorising}. Such datasets have the benefit that it is possible to alter the parameters or \emph{factors} of the data-generating process explicitly and measure directly the effect this has on the model's internal representations. This limits the application of such metrics, as direct access to the ground-truth data-generating factors of real-world datasets is impossible. For real-world datasets, these values can only be approximated by extensive annotation of samples with some chosen set of semantically descriptive attributes - e.g. annotating images of humans with information about age, gender, background type and so on.

The second limitation is the assumption of a probabilistic generative model, typically some form of Variational Auto-Encoder. Such models consist of a probabilistic encoder learning a latent space representation $\mathbf{z}$ of the input data $\mathbf{x}$ by approximating the distribution $p(\mathbf{z}|\mathbf{x})$ and a decoder parameterising $q(\mathbf{x}|\mathbf{z})$. Many disentanglement metrics, such as those proposed in \citet{udr_disent_metric_vaes} and \citet{unsupervised_disent_metrics} are concerned with measures of mutual information and conditional entropy between different signal components. While these measures are informative for probabilistic models, they are provably vacuous for deterministic NNs such as standard Convolutional NNs and Transformers. As demonstrated in \citet{mutual_information_in_deterministic_nns}, The conditional entropy $H(Z|X)$ is no longer meaningful in the information-theoretic sense when $Z$ is a deterministic function of $X$.

The task of quantitatively assessing signal disentanglement in deterministic models without access to the ground-truth data generation process, therefore, limits the set of available metrics. However, \citet{content_style_disent_metric} propose the use of two metrics to measure the disentanglement of the representations of style and content in an image, which bears some similarities to our goal of quantifying the disentanglement of causal and confounder signals relative to some input data. The first of these two metrics is Distance Correlation ($DC$). Proposed in \citep{distance_correlation_original_paper}, $DC$ is a well-established measure of the dependence between two variables. The second is Information Over Bias ($IoB$), proposed in \citet{content_style_disent_metric}, which uses the reconstruction error of a NN trained to reconstruct a signal $\mathbf{x}$ from a representation $\mathbf{z}$ as a measure of the information content of $\mathbf{z}$ with respect to $\mathbf{x}$. 

\subsubsection{Distance Correlation}
Given a set of $N$ pairs of vector or tensor-valued samples $\{(\mathbf{u}, \mathbf{v})\}_{n=1}^N = (\mathbf{U}, \mathbf{V})$, the $DC$ is defined as follows. Let $\mathbf{A}^*$ and $\mathbf{B}^*$ be the unnormalised distance matrices of $\mathbf{u}$ and $\mathbf{v}$ respectively, under some distance metric $||\ .\ ||$, such that $\mathbf{A}^*_{i,j} = ||\mathbf{u}_i - \mathbf{u}_j||$ and $\mathbf{B}^*$ is defined similarly for $\mathbf{v}$. A normalisation is then applied by subtracting off the column-mean and the row-mean and adding the global mean of each matrix to obtain $\mathbf{A}$ and $\mathbf{B}$ where $\mathbf{A}_{i,j} = \mathbf{A}^*_{i,j} - \bar{\mathbf{A}}_{i,.} - \bar{\mathbf{A}}_{.,j} + \bar{\mathbf{A}}_{.,.}$. The squared distance covariance $dCov$ is defined as the arithmetic mean of $\mathbf{A}_{i,j}\mathbf{B}_{i,j}$ over all the $N$ samples. The $DC$ is then calculated analogously to a correlation coefficient, as the normalised distance covariance: 

\begin{align*}
    \begin{split}
        DC(\mathbf{U}, \mathbf{V}) &= \frac{dCov(\mathbf{U},\mathbf{V})}{\sqrt{dCov(\mathbf{U},\mathbf{U})dCov(\mathbf{V},\mathbf{V})}},\\
        dCov(\mathbf{U},\mathbf{V}) &= \sqrt{\sum_{i=1}^N\sum_{j=1}^N\frac{\mathbf{A}_{i,j}\mathbf{B}_{i,j}}{N^2}}.
    \end{split}
\end{align*}

Unlike Pearson's Correlation Coefficient, a value of $DC(X, Y) = 0$ implies that $X$ and $Y$ are independent. Note also that $DC$ allows for the measurement of dependence between variables of different dimensionalities. The computation of $DC(X, Y)$ requires only that a distance metric is defined between samples of the \emph{same} variable $||x_i - x_j||$, but crucially does \emph{not} require $||x_i - y_i||$ to be defined. Importantly for our application, this allows us to compute the dependence between e.g. a vector $\mathbf{c}$ and a $channel \times width \times height$ image tensor $\mathbf{x}$. $DC$ is therefore a general measure of the dependence between two variables.

\subsubsection{Information over Bias}
Given some input data $\mathbf{x}$ and a learned representation $\mathbf{z}$, a decoder network $g_\theta$ is trained to reconstruct $\mathbf{x}$ from $\mathbf{z}$. The $IoB$ is then defined as the average reconstruction performance gain, in terms of the Mean Squared Error, when operating on $\mathbf{z}$ compared to on  $\mathbf{1}$, a dummy input vector of ones:

\begin{equation}
    IoB(\mathbf{x}, \mathbf{z}) = 
    \frac{1}{N}\sum_{i=1}^N 
        \frac
        {MSE(\mathbf{x}_i, g_\theta(\mathbf{1}))}
        {MSE(\mathbf{x}_i, g_\theta(\mathbf{z}_i))}.
    \label{eq:iob}
\end{equation}

Like $DC$, $IoB$ is attractive as a metric because it admits both tensor and vector representations for the signals $\mathbf{x}$ and $\mathbf{z}$, and does not require $\mathbf{x}$ and $\mathbf{z}$ to be of the same size or dimensionality. It offers a flexible measure of relative information content, without being restricted to stochastic signals.

\section{Related Work}\label{sec:related_work}
For completeness, we briefly review a few other causal models and explain why they are not studied in this paper, followed by an overview of related approaches for measuring model disentanglement.
\subsection{Other Causal Neural Network Models}
In this paper, we are concerned with models which aim to explicitly model causal and confounder signals, with the goal of using the causal signal for robust predictions. Approaches such as \citet{vit_causal_intervention}, where Causal Inference is successfully used to create heuristic metrics for the detection of adversarial attacks also exist. This approach uses Causal Inference to motivate the analysis of the model but does so as a second step on top of the trained model, and therefore falls outside the model type considered in this paper. Models from domains other than image recognition are also of relevance, although outside the scope of this paper. In \citet{causal_intervention_nlp}, the Natural Language Processing model uses latent-space smoothing over the confounder signal in a similar manner to CausalAdv to increase adversarial robustness.

CATT, proposed in \citet{CATT}, has a similar design philosophy to the models investigated in this paper, although the causal intervention is performed as a front-door adjustment. However, the marginalisation over the confounding signal is absorbed into the model's intersample and intrasample attention mechanisms. This obfuscates the measurement of the $C$ and $S$ signals without the application of additional modelling assumptions. CONTA, as proposed in \citet{CONTA_semantic_segmentation}, is another related model, where the confounder signal is not constructed on a per-instance basis as in the models presented here, but rather as an average pixel classification importance map across all samples in a class. However, both CONTA and CATT could be interesting objects of future work in the measurement and analysis of causal disentanglement.

\subsection{Disentanglement of Representations}
In terms of measuring the disentanglement of different model architectures, \citet{challenging_common_assumptions_disentangled_representations} offer a thorough investigation of VAE-style models on the task of learning disentangled representations in an unsupervised fashion for seven synthetic datasets. Similarly, \citet{evaluating_disentangled_representations} investigate the performance of disentanglement metrics for VAE models on synthetic datasets and propose a new quantitative metric for measuring this disentanglement.

In contrast, we focus on measuring the disentanglement of Causal NN models with metrics which are generally applicable also to deterministic models trained on real-world datasets without access to the true data-generating factors. The most relevant paper to this end is probably \citet{content_style_disent_metric} which aims to measure the disentanglement of content and style in three representative computer vision models. However, this is not in the context of causal disentanglement nor is it related to adversarial robustness. 

\section{Methodology}
In this section, we detail the motivation for and setup of the experiments conducted, as well as the choice of causal and confounder signals for each model. With these experiments, we specifically aimed to address the following research questions.

\paragraph{RQ1:}
To what extent and in what way do the investigated models exhibit causal disentanglement?
\paragraph{RQ2:}
What is the relationship between the measured metric values and the models' performance?
\paragraph{RQ3:}
What is the relationship between the measured metric values and the models' robustness to adversarial attacks?

\subsection{Measurements}
As the models were trained on real-world datasets without any other annotation than class labels, the choice of exactly which aspects of the models' signals to measure does not have a unique well-defined answer a priori. Therefore, we selected five measurements which we believe each capture important aspects of the models' causal disentanglement behaviour. These measurements are variations on the ones proposed in \citet{content_style_disent_metric} and are defined and motivated in this subsection as well as summarised in Table \ref{tab:measurements}.

\subsubsection{Separation of Causal and Confounder Signals}
Perhaps the most central characteristic of the signal flow in Causal NNs is the separation of the signal streams of the causal signal $\mathbf{c}$ and the confounder signal $\mathbf{s}$. The way we chose to quantify this behaviour was by measuring the $DC$ between these two signal streams. A high $DC(C, S)$ means that $C$ and $S$ are correlated and dependent, which is contrary to the goal of Causal NNs. We, therefore, take a high $DC(C, S)$ value to indicate low causal disentanglement. The first measurement is then defined as $M_1 = 1 - DC(C, S)$, so that a high value of $M_1$ corresponds to a high degree of causal/confounder separation.

\subsubsection{Causal Signal Informativeness}
Since the Causal NNs studied in this work by definition employ the causal signal $\mathbf{c}$ in performing their primary task, we believe it is useful to measure the information content of this signal with respect to the input $\mathbf{x}$. In our experiments, this was done with two separate measurements. The first is $M_2 = DC(X, C)$ which measures the correlation between the causal signal and the input image. The second measurement is based on $IoB(X, C)$, that is how well the input image $\mathbf{x}$ can be reconstructed on a pixel level from $\mathbf{c}$ relative to from an uninformative signal. $IoB(X, C)$ takes on its minimum value of $1$ when the causal signal is completely uninformative, and higher values indicate higher informativeness. To normalise the range of our measurements we reciprocate the ratio and define $BoI(X, C) = \frac{1}{IoB(X, C)}$, and let $M_4 = 1 - BoI(X, C)$. $M_4$ is now in the range $[0, 1]$ and higher values indicate higher pixel-level information content of $\mathbf{c}$ with respect to $\mathbf{x}$.

\subsubsection{Confounder Signal Informativeness}
What the desirable properties of the confounder signal $\mathbf{s}$ are in Causal NNs is still an open research question. It is argued by \citet{content_style_disent_metric} that when measuring content/style disentanglement it is necessary for the style signal to be informative with respect to the input image. This is because a style signal consisting of e.g. random noise would be disentangled from the content signal in the sense that the two would be independent. \citet{content_style_disent_metric} consider this a failure mode of their content/style disentanglement and argue that in order to rule out such failure an informative style signal is necessary. Our experiments are concerned with the disentangling of causal and confounder signals, and we believe it is not a priori obvious which properties of the confounder signal $\mathbf{s}$ are beneficial to the performance and robustness of Causal NNs. Nonetheless, we believe that the semantic information that Causal NNs encourage in the confounder signal stream, such as information about background, lighting, and camera angle, bears similarities with the information intended for the style signal in content/style disentangled NNs. Hence, we define the measurement $M_3 = DC(X, S)$ to assess the dependency between the input image $\mathbf{x}$ and the confounder signal $\mathbf{s}$. Similarly to $M_4$ we finally define $M_5 = 1 - BoI(X, S)$ to measure the pixel-level reconstructive information in the confounder signal.

\begin{table}
    \centering
    \caption{The measurements taken of the models investigated. All measurement values are in the range $[0,1]$.}
    \begin{tabular}{c|l|l}
        \toprule
        $M_i$ & \textbf{Value} & \textbf{Interpretation} \\
        \midrule
        $M_1$ & $1-DC(C,S)$ & Causal/confounder separation\\
        \hline
        $M_2$ & $DC(X,C)$ & Input/causal signal correlation\\
        \hline
        $M_3$ & $DC(X,S)$ & Input/confounder signal correl.\\
        \hline
        $M_4$ & $1-BoI(X,C)$ & Pixel info in causal signal\\
        \hline
        $M_5$ & $1-BoI(X,S)$ & Pixel info in confounder signal\\
        \bottomrule
    \end{tabular}
    \label{tab:measurements}
\end{table}

\subsection{Model Selection}
The four models selected for analysis in this paper have shown good performance on challenging primary tasks such as AA robustness and rare-context image classification. We have chosen to study the disentanglement behaviour of these models because they all explicitly aim to separate the modelling of causal and spurious signals, and argue that this causal consistency is the reason for each model's high performance. The models were published in the period 2020 to 2022 and we believe they are representative of the current state-of-the-art in Causal NN models.

\subsection{Choice of Causal and Confounder Signals}
Throughout our analysis, the causal and confounder signals for each model were taken as follows:
\subsubsection{CAMA}
The value of $S$ is sampled once per input image from the latent style representation of the final encoder network as $S \sim q(S|X, Y, M)$, and $C$ is taken as the hidden-state representation $\mathbf{h}_y$ of the label $\mathbf{y}$ as produced by the pre-merge step in the decoder.
\subsubsection{CaaM}
$C$ and $S$ were taken as the outputs $\mathbf{c}$ and $\mathbf{s}$ of the final disentanglement block of the CNN-CaaM model with a ResNet18 backbone.
\subsubsection{CausalAdv}
After the latent space embedding of the input as $\mathbf{h} = \text{ResNet18}(\mathbf{x})$, $C$ was taken as the projection $\mathbf{c} = \mathbf{W}_c\mathbf{h}$. $S$ was chosen as $\mathbf{s} = \mathbf{W}_s\mathbf{h}$, i.e. before the addition of the gaussian noise $\mathbf{n}$.
\subsubsection{DICE}
For DICE, S was taken as the embedded confounder sample $\mathbf{s} = \text{ResNet18}(\mathbf{s}_x)$ and C as the embedding of $\mathbf{x}_c$, i.e. the input image $\mathbf{x}$ after the backdoor-adjustment of Equation \ref{eq:dice_backdoor} has been approximated as $\mathbf{x}_c = \mathbf{x} + \sum_{\mathbf{s}\in \mathbb{S}}P(\mathbf{s})\mathbf{s}$

\subsection{Experimental Setup}
The four models we have studied vary in terms of their intended use case, as well as their natural performance on their primary tasks. In order to make as fair a comparison as possible we altered or re-implemented DICE, CaaM, and CausalAdv to employ the same ResNet18 backbone architecture. CAMA, being structured as a VAE, differs quite significantly from the other three and does not rely on the same type of initial input data latent-space embedding. In order to not deviate too much from CAMA's original design, we opted to keep the architecture as described in \citet{CAMA}. We conducted all experiments using the three standard image recognition benchmarking datasets MNIST \citep{mnist_original_paper}, CIFAR10, and CIFAR100 \citep{cifar_original_paper}. All models were trained for a fixed number of epochs, and the model with the highest validation accuracy on the clean dataset was returned for testing in each case. For all datasets, the original training split was randomly partitioned into train and validation splits in the ratio $4:1$.

\subsubsection{Metrics and Measurements}
All $DC$ values were computed over each dataset's test split, and $IoB$ models were trained on the train split and tested on the test split. The training budget for each model was set to roughly match the training setup in the respective original papers. For the training of decoder models in the computation of $IoB$, $20\%$ of the available training data was randomly selected as a validation split, and models returned when no validation improvements were seen for $40$ epochs. During the tracking of disentanglement metrics throughout entire training runs, this validation patience was lowered to $5$ epochs and the total training budget was capped at $50$ epochs.

\subsubsection{Adversarial Robustness}

\begin{table}[h]
    \centering
    \caption{Adversarial attacks used to test the robustness of models, with the number of iteration steps, maximum perturbations, and distance norms.}
    \begin{tabular}{c|c|c|c}
        \toprule
        Attack & \# Steps & Norm & $\epsilon$ \\
        \midrule
        PGD & 20 & $l_2$ & $1.0$\\
        \hline
        PGD & 40 & $l_2$ & $1.0$\\
        \hline
        PGD & 20 & $l_\infty$ & $\frac{8}{255}$\\
        \hline
        PGD & 40 & $l_\infty$ & $\frac{8}{255}$\\
        \hline
        FGSM & - &  $l_\infty$ & $\frac{8}{255}$\\
        \hline
        CW & 20 & $l_2$ & $1.0$\\
        \hline
        CW & 40 & $l_2$ & $1.0$\\
        \bottomrule
    \end{tabular}
    \label{tab:attack_specs}
\end{table}

In order to assess the robustness of the models we used the three standard attack algorithms $PGD$, $FGSM$, and $CW$ under different distance norms and optimisation budgets for a total of 7 attack configurations - these are enumerated in Table \ref{tab:attack_specs}. 

When measuring robustness we first measured the models' classification accuracy on the unperturbed test split of each dataset to get the clean accuracy $a_c$. We then attacked each dataset's test split with each of the seven attack configurations and measured the models' resulting perturbed accuracy $a_p$. Finally, we calculated the absolute performance drop as $\Delta_{abs} = a_c - a_p$ and the relative performance drop as $\Delta_{rel} = \frac{\Delta_{abs}}{a_c}$.

\section{Results and Analysis}
In this section, we present the results of our experimental evaluation of the four chosen models, as well as analyse and discuss the findings in light of our three research questions.

\subsection{RQ1: Observed Disentanglement}
\begin{table*}
    \centering
    \caption{Values of the five measured metrics, averaged over the three datasets for each model. Values are given as mean $\pm$ std.}
    \begin{tabular}{c|cc|cc|cc|cc|cc}
        \toprule
        Model & 
        \multicolumn{2}{|c|}{$1 - DC(C, S)$} & 
        \multicolumn{2}{|c|}{$DC(X, C)$} & 
        \multicolumn{2}{|c|}{$DC(X, S)$} & 
        \multicolumn{2}{|c|}{$1 - BoI(X, C)$} & 
        \multicolumn{2}{|c}{$1 - BoI(X, S)$} \\
        \midrule
        CAMA & 
        \multicolumn{2}{|c|}{$\mathbf{0.917} \pm 0.01$} & 
        \multicolumn{2}{|c|}{$0.442 \pm 0.21$} & 
        \multicolumn{2}{|c|}{$\mathbf{0.711} \pm 0.18$} & 
        \multicolumn{2}{|c|}{$0.156 \pm 0.10$} & 
        \multicolumn{2}{|c}{$0.368 \pm 0.22$} \\
        \hline
        CaaM & 
        \multicolumn{2}{|c|}{$0.132 \pm 0.16$} & 
        \multicolumn{2}{|c|}{$0.512 \pm 0.16$} & 
        \multicolumn{2}{|c|}{$0.493 \pm 0.18$} & 
        \multicolumn{2}{|c|}{$0.415 \pm 0.08$} & 
        \multicolumn{2}{|c}{$\mathbf{0.460} \pm 0.06$} \\
        \hline
        CausalAdv & 
        \multicolumn{2}{|c|}{$0.819 \pm 0.09$} & 
        \multicolumn{2}{|c|}{$0.524 \pm 0.14$} & 
        \multicolumn{2}{|c|}{$0.128 \pm 0.07$} & 
        \multicolumn{2}{|c|}{$0.190 \pm 0.12$} & 
        \multicolumn{2}{|c}{$0.006 \pm 0.01$} \\
        \hline
        DICE & 
        \multicolumn{2}{|c|}{$0.424 \pm 0.28$} & 
        \multicolumn{2}{|c|}{$\mathbf{0.634} \pm 0.09$} & 
        \multicolumn{2}{|c|}{$0.587 \pm 0.25$} & 
        \multicolumn{2}{|c|}{$\mathbf{0.455} \pm 0.03$} & 
        \multicolumn{2}{|c}{$0.275 \pm 0.19$} \\
        \bottomrule
    \end{tabular}
    \label{tab:avg_metrics}
\end{table*}

The full set of measurement values across each of the tested models is shown in Table \ref{tab:avg_metrics}. The first thing to note is that although all models aim to disentangle the causal and confounder signal streams, there is a large variation in how well $C$ and $S$ are decorrelated. The VAE-style model CAMA achieves the highest separation with an average value of $M_1 = 1 - DC(C, S)$ of 0.917, close to full statistical independence between $C$ and $S$. The lowest level of decorrelation is achieved by CaaM with an average $M_1$ value of 0.132. All models score on average 0.442 or higher on the correlation of causal signal and input content as measured by $M_2 = DC(X, C)$. This is to be expected as the causal signal stream $\mathbf{c}$ is used by each model to make classification predictions, and hence a high $M_2$ value is directly encouraged during training.

The models vary considerably in terms of how correlated the confounder signal and input are, from CAMA with an average $M_3 = DC(X, S)$ of 0.711 to CausalAdv with a value of 0.128. We see that CausalAdv \emph{consistently} exhibits low correlation between confounder and input across all three datasets with a standard deviation of only 0.07. Note that the confounder signal is measured \emph{before} the addition of the Gaussian noise term in this model, which makes the low value even more notable. In terms of pixel-level information, it is interesting to note that even though CAMA is a VAE-type model and aims to reduce reconstruction loss during training, this is not the model which best manages to reconstruct the input from either the causal or confounder signal. Finally, we note that DICE's causal signal is both the most correlated with the input signal, and the causal signal which is best able to reconstruct the input, indicating high causal signal information content. Similarly, CausalAdv's confounder signal is both the least correlated with the input and has the least capacity to reconstruct the input.

\subsubsection*{Summary} Even though all models aim to disentangle the causal and confounder signal streams, there is a large variation in the extent to which these signal streams are decorrelated as measured by $DC(C, S)$. There is also moderate variation between the models in terms of the information content of the confounder stream with respect to the input as measured both by the $DC(X, S)$ and $1 - BoI(X, S)$.

\subsection{RQ2: Disentanglement and performance}\label{sec:rq2}
\begin{figure}[h]
    \centering
    \includegraphics[width=0.90\linewidth]{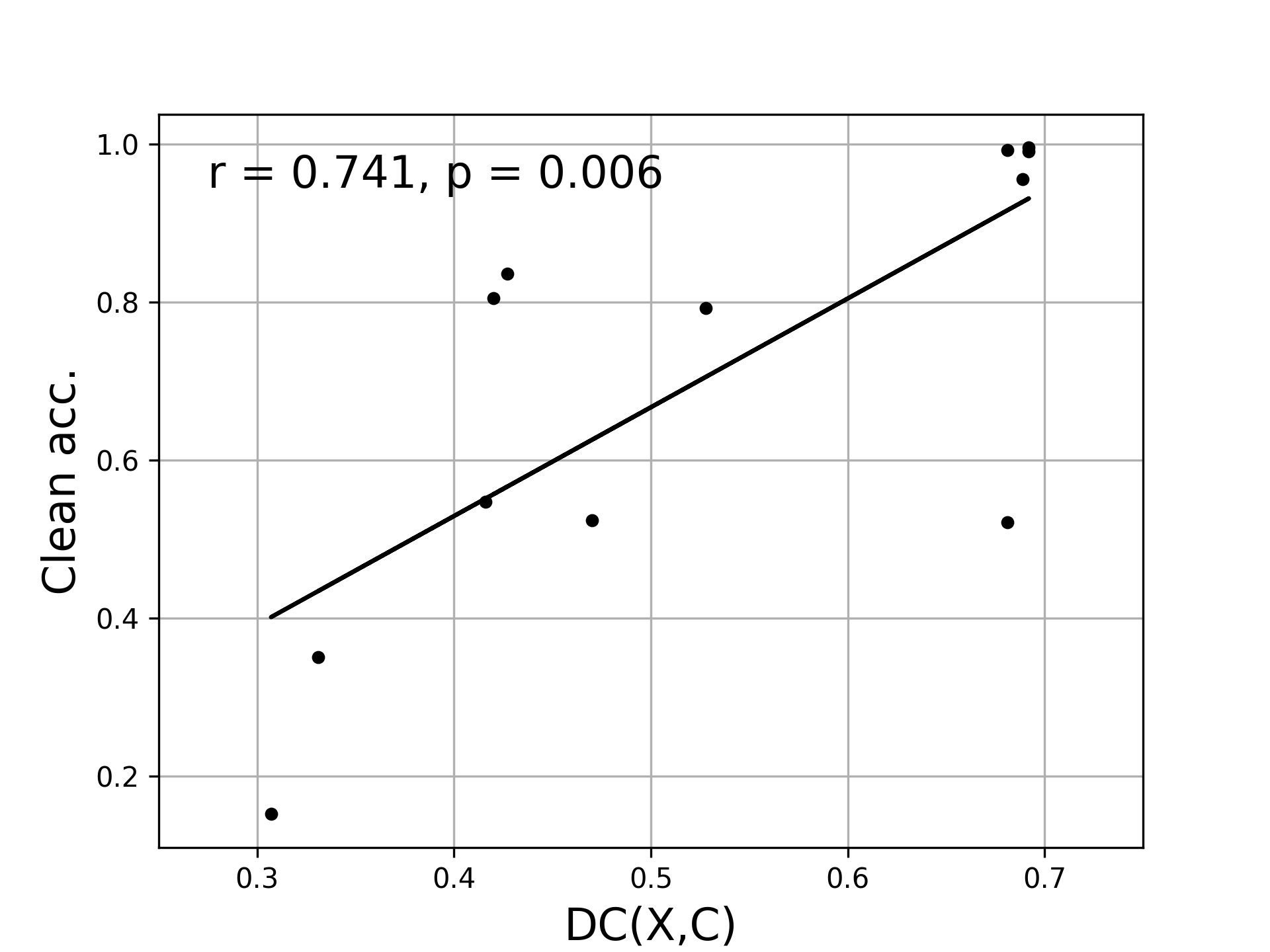}
    \caption{$DC(X, C)$ vs clean test classification accuracy for all models on all datasets. Linear best-fit line in black.}
    \label{fig:DCXC_vs_clean_acc}
\end{figure}

The only measurement value with a statistically significant correlation with a model's performance on its primary task is $M_2 = DC(X, C)$, the distance correlation between the causal signal stream and the input image. The Pearson Correlation Coefficient (PCC) between $M_2$ and a model's clean test classification accuracy $a_c$ is $r=0.741$ at a p-value of $p=0.006$. This relationship is also illustrated in Figure \ref{fig:DCXC_vs_clean_acc} which plots $DC(X, C)$ values vs clean test accuracy for all models on all datasets.

\begin{figure}[h]
  \centering
  \includegraphics[width=0.90\linewidth]{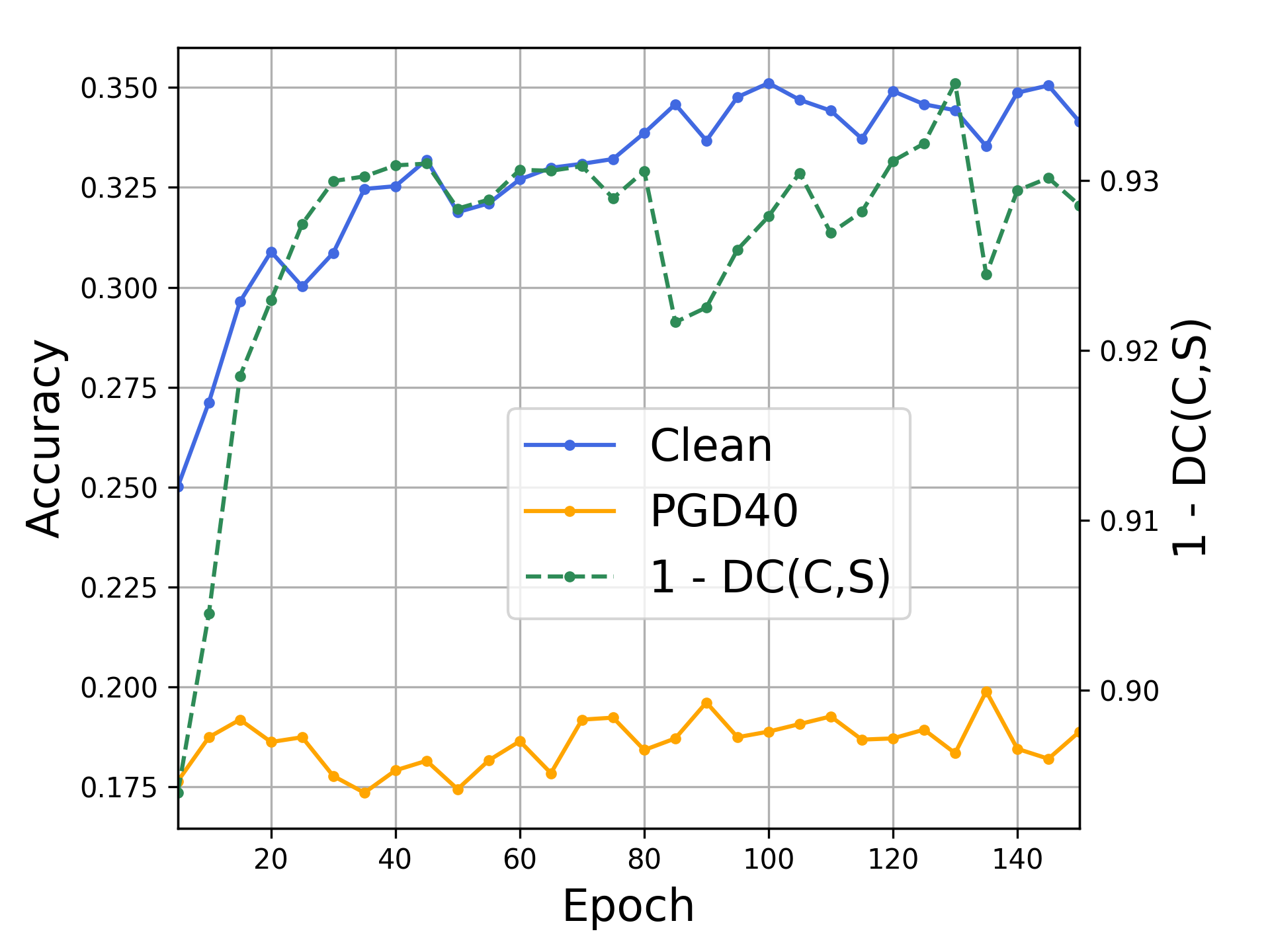}
  \includegraphics[width=0.90\linewidth]{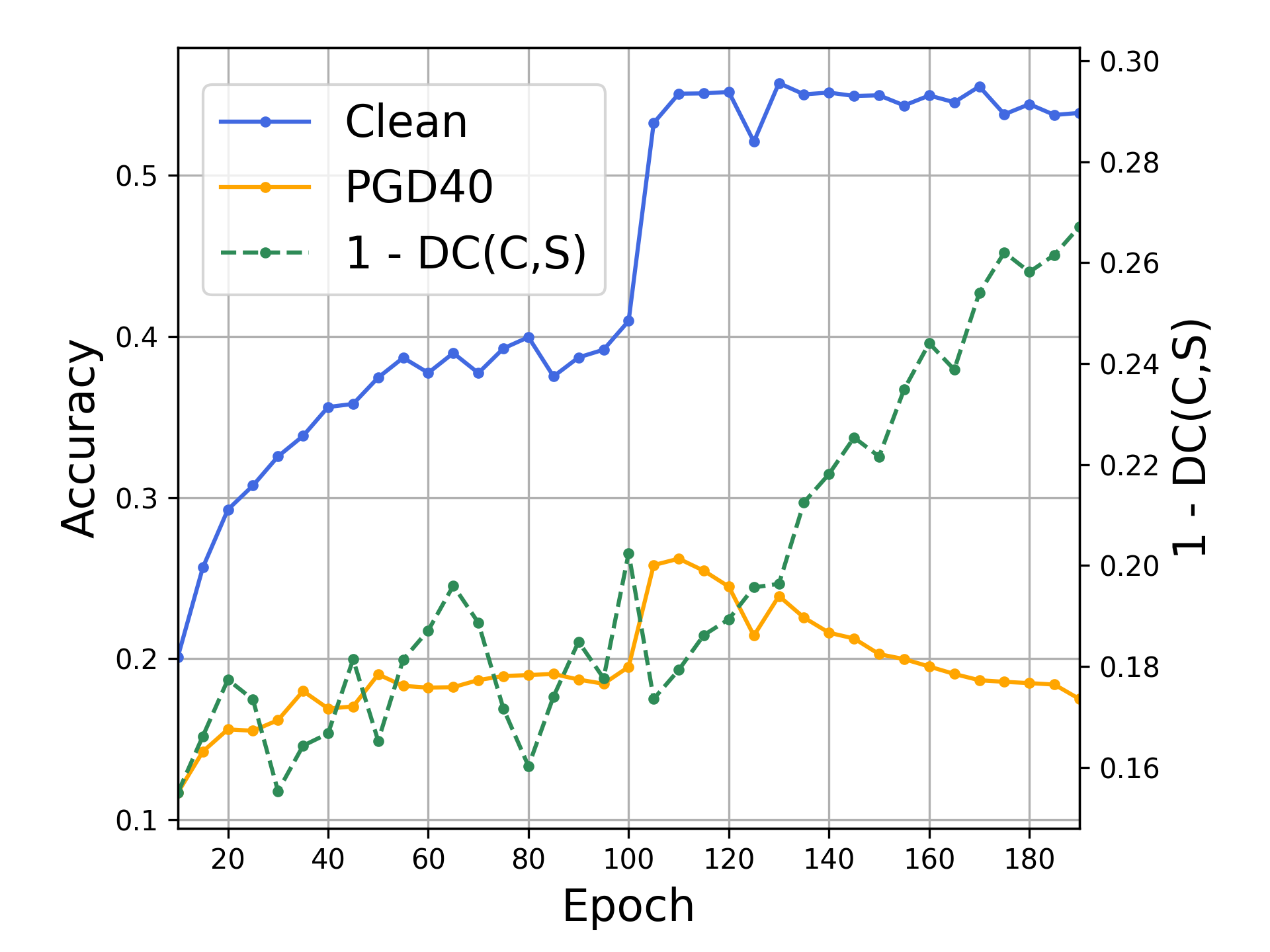}
  \caption{$1 - DC(C, S)$, clean and adversarial accuracy as a function of training epoch for CAMA on CIFAR10 (top), and DICE on CIFAR100 (bottom).}
  \label{fig:m1_vs_epoch}
\end{figure}

The green dashed line in Figure \ref{fig:m1_vs_epoch} shows the evolution of measurement $M_1 = 1 - DC(C, S)$ as a function of training epoch for CAMA training on CIFAR10 and DICE training on CIFAR100. We note that both models see a decrease in $M_1$ as training progresses, which corresponds to the type of disentanglement encouraged by both models' inductive biases.

\subsubsection*{Summary}
There is a strong association ($r=0.741$) between the distance correlation of the causal signal and the input and the clean test accuracy of a model. Apart from this, no measurement shows a statistically significant association with model performance on clean data.

\subsection{RQ3: Disentanglement and Robustness}

\begin{table}
    \centering
    \caption{Clean accuracy, average adversarial accuracy, and relative accuracy drop $\Delta_{rel}$ for all tested models and datasets.}
    \begin{tabular}{c|c|c|c|c}
        \toprule
        Dataset & Model & Clean & Mean $a_p$ & $\Delta_{rel}$ \\
        \midrule
        
        \multirow{4}{*}[0em]{MNIST} 
        & CAMA & 95.6\% & 81.8\% & 14.5\% \\
        \cline{2-5}
        {} & CaaM & \textbf{99.6\%} & 18.3\% & 81.7\% \\
        \cline{2-5}
        {} & CausalAdv & 99.3\% & \textbf{97.4\%} & \textbf{1.9\%} \\
        \cline{2-5}
        {} & DICE & 99.1\% & 97.0\% & 2.1\% \\
        
        \hline

        \multirow{4}{*}[0em]{CIFAR10} 
        & CAMA & 35.1\% & 23.3\% & \textbf{33.6\%} \\
        \cline{2-5}
        {} & CaaM & \textbf{83.6\%} & 4.5\% & 94.6\% \\
        \cline{2-5}
        {} & CausalAdv & 80.5\% & \textbf{45.4\%} & 43.6\% \\
        \cline{2-5}
        {} & DICE & 79.3\% & 39.7\% & 49.9\% \\
        
        \hline

        \multirow{4}{*}[0em]{CIFAR100} 
        & CAMA & 15.2\% & 9.2\% & \textbf{39.3\%} \\
        \cline{2-5}
        {} & CaaM & \textbf{54.7\%} & 1.7\% & 96.8\% \\
        \cline{2-5}
        {} & CausalAdv & 52.4\% & \textbf{23.7\%} & 54.8\% \\
        \cline{2-5}
        {} & DICE & 52.1\% & 22.3\% & 57.2\% \\

        \bottomrule
    
    \end{tabular}
    \label{tab:dccs_vs_robustness}
\end{table}

Table \ref{tab:dccs_vs_robustness} shows the clean and adversarial accuracy of all models on all datasets, as well as the relative adversarial performance decrease $\Delta_{rel}$. There is some variation in the clean data performance between models, with CaaM achieving the highest accuracy for all datasets. CAMA scores significantly lower than the other models on both CIFAR10 and CIFAR100, but these accuracies are within expectations for a simple VAE-style model. CausalAdv and DICE achieve the best and second-best average adversarial accuracies respectively, which is also reasonable given that these two models use adversarial training with PGD10 attacks as part of their training loops. More surprising is the relative robustness of CAMA, which only uses slight rotations and translation of input images to model adversarial perturbations during training. Finally, we observe that CaaM suffers the largest performance degradation under adversarial attacks, by a large margin.

\begin{table*}
    \centering
    \caption{Pearson Correlation Coefficients with p-values of the five metrics with clean accuracy and average adversarial performance degradation across all attacks. Results significant at p=0.05 highlighted in \textbf{bold}.}
    \begin{tabular}{c|cc|cc|cc|cc|cc}
        \toprule
         & 
        \multicolumn{2}{|r|}{$1 - DC(C, S)$} & 
        \multicolumn{2}{|c|}{$DC(X, C)$} & 
        \multicolumn{2}{|c|}{$DC(X, S)$} & 
        \multicolumn{2}{|c|}{$1 - BoI(X, C)$} & 
        \multicolumn{2}{|c}{$1 - BoI(X, S)$}\\
        \midrule
        Clean acc. & 
        \multicolumn{2}{|r|}{$-0.275\ p>.050$} & 
        \multicolumn{2}{|r|}{$\mathbf{0.741}\ p=.006$} & 
        \multicolumn{2}{|r|}{$-0.410\ p>.050$} & 
        \multicolumn{2}{|r|}{$ 0.299\ p>.050$} & 
        \multicolumn{2}{|r}{$ -0.430\ p>.050$}\\
        \hline
        Adv acc. & 
        \multicolumn{2}{|r|}{$ 0.429\ p>.050$} & 
        \multicolumn{2}{|r|}{$ \mathbf{0.638}\ p=.026$} & 
        \multicolumn{2}{|r|}{$-0.377\ p>.050$} & 
        \multicolumn{2}{|r|}{$-0.189\ p>.050$} & 
        \multicolumn{2}{|r}{$ \mathbf{-0.725}\ p=.008$}\\
        \hline
        $\Delta_{abs}$ & 
        \multicolumn{2}{|r|}{$\mathbf{-0.820}\ p=.001$} & 
        \multicolumn{2}{|r|}{$-0.048\ p>.050$} & 
        \multicolumn{2}{|r|}{$ 0.056\ p>.050$} & 
        \multicolumn{2}{|r|}{$ 0.543\ p>.050$} & 
        \multicolumn{2}{|r}{$  0.476\ p>.050$}\\
        \hline
        $\Delta_{rel}$ & 
        \multicolumn{2}{|r|}{$\mathbf{-0.720}\ p=.008$} & 
        \multicolumn{2}{|r|}{$-0.343\ p>.050$} & 
        \multicolumn{2}{|r|}{$ 0.135\ p>.050$} & 
        \multicolumn{2}{|r|}{$ 0.437\ p>.050$} & 
        \multicolumn{2}{|r}{$ \mathbf{0.597}\ p=.040$}\\
        \bottomrule
    \end{tabular}
    \label{tab:pcc_of_measurements}
\end{table*}

In order to assess the association between the different measurements made and model robustness quantitatively, Table \ref{tab:pcc_of_measurements} shows the PCC of the five measurements taken for each model and each model's clean accuracy, average adversarial accuracy across the seven attacks used, and corresponding average absolute and relative performance drop. At a significance threshold of $p=0.05$ there are five statistically significant associations.

Firstly we see that a high $M_2 = DC(X, C)$ value is associated with both a high clean test accuracy (see Section \ref{sec:rq2}) and a high average adversarial accuracy ($r=0.638, p=0.026$). This is likely because the causal signal $\mathbf{c}$ is used directly for classification, and hence a higher correlation with the input image makes the model's prediction task easier. It is interesting to see that high \emph{pixel-level} information content in the casual signal as measured by $M_4 = 1 - BoI(X, C)$ is \emph{not} associated with either clean or adversarial accuracy. This could indicate that the information in the causal signal should capture more high-level features of the input rather than low-level pixel information in order for the model to make accurate predictions.

We also see that high pixel-level information in the confounder signal $\mathbf{s}$ in terms of $1 - BoI(X, S)$ is moderately associated with relative adversarial performance degradation (r=0.597, p=0.040), although this association is no longer significant when accuracy degradation is measured in absolute terms. This gives some indication that low-level input information in the confounder signal hurts model robustness. This is interesting as it goes against what is argued by \citet{content_style_disent_metric}, namely that pixel-level informative content and style signals are desirable disentanglement properties. However, this is in line with the recent trend of encouraging higher-level semantic content rather than low-level pixel information in learned representations as seen in e.g. \citet{lecun2022path}.

\begin{figure}[h]
    \centering
    \includegraphics[width=0.90\linewidth]{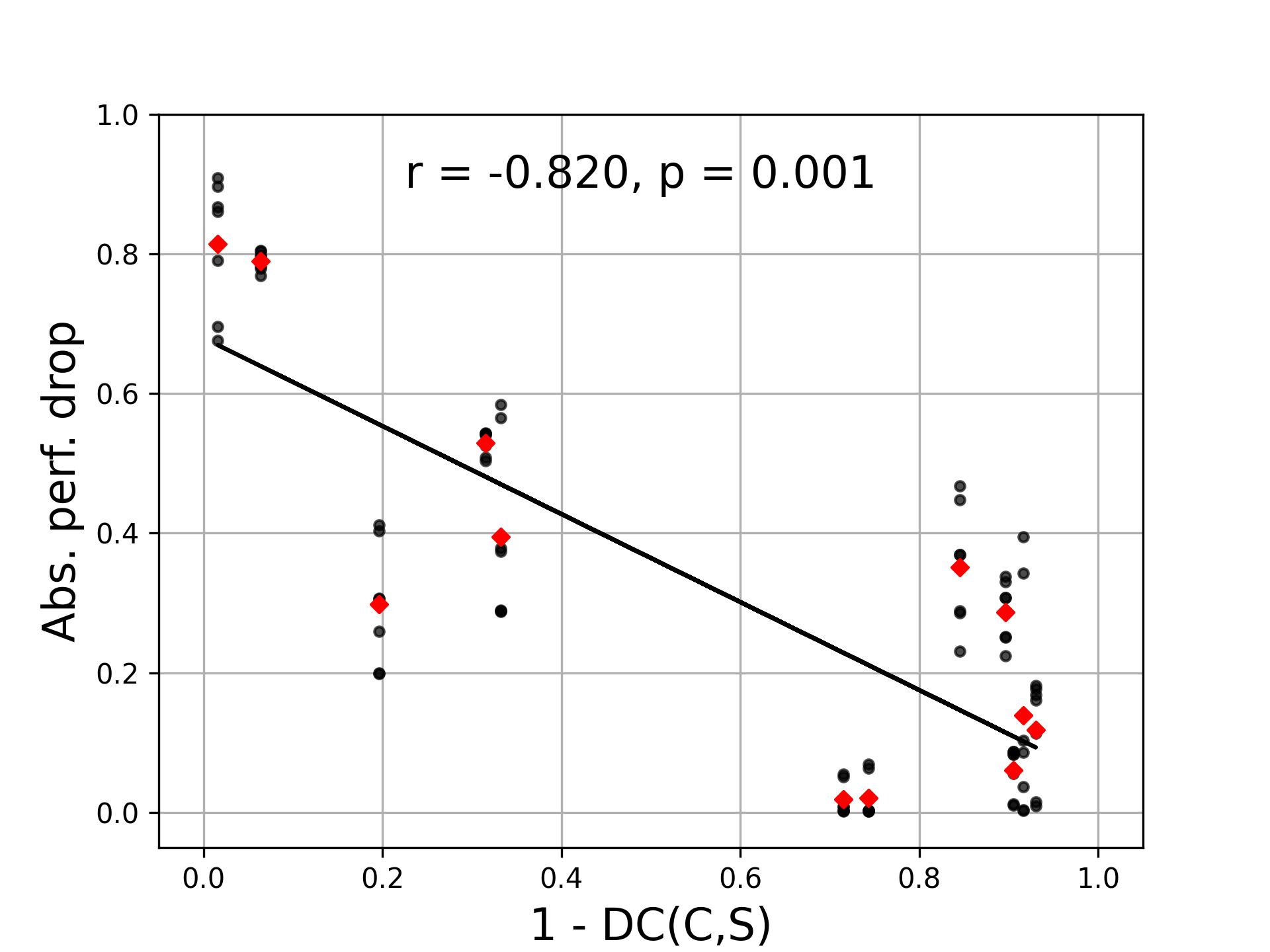}
    \caption{$1 - DC(C, S)$ vs absolute performance drop for all tested models and datasets. Mean performance drop indicated by red diamonds.}
    \label{fig:DC_vs_perf_drop}
\end{figure}

The strongest correlations we find are between the decorrelation of the causal and confounder signal $M_1 = 1 - DC(C, S)$ and adversarial robustness. Decorrelation is strongly negatively associated with both absolute (r=-0.820, p=0.001) and relative (r=-0.720, p=0.008) adversarial performance drop. This is strong evidence in support of the notion that causally disentangled representations are beneficial for adversarial robustness. This relationship is also illustrated in Figure \ref{fig:DC_vs_perf_drop} which shows the value of $M_1$ against $\Delta_{abs}$ for all models, datasets and attacks in black, with average performance drop across all attacks indicated by red diamonds. However, it is interesting to note that the bottom plot in Figure \ref{fig:m1_vs_epoch} shows a point during model training after which $M_1$ increases and adversarial accuracy under the PGD40 attack decreases. This could indicate that there is a sweet spot during model training, after which the increasing $M_1$ is a result of model overfitting.

\subsubsection*{Summary}
We observe a strong association between the decorrelation of causal and confounder signals and a model's adversarial robustness ($r=0.820, p=0.001$). This supports the idea that causal disentanglement helps robustness.

\section{Key Findings and Conclusions}
In this paper, we investigated the causal disentanglement of four state-of-the-art Causal NN models. We used metrics from content/style disentanglement to assess different aspects of the separation and information content of the causal and confounder signals in each model without requiring access to the ground-truth data-generating function or restricting our analysis to stochastic models. Finally, we quantitatively assessed the association between the metrics and both the clean performance and the adversarial robustness of the models under a range of different common attacks.

\subsection{Key Findings}
\begin{itemize}
    \item Although each model aims to separate the representations of causal and confounder signals, there is a large variation in how well this aim is achieved.
    \item High distance correlation between the causal and input signals is associated with higher classification accuracy on both clean and adversarially perturbed test data.
    \item The decorrelation of causal and confounder signals is strongly associated with adversarial robustness.
\end{itemize}

\subsection{Conclusions}
Our findings point in the direction that the decorrelation of causal and confounder signals is useful for achieving robust Causal NNs, whereas low-level pixel information content appears at least unhelpful for the causal signal and seems to degrade robustness in the confounder stream. This indicates that the appropriate  signal decorrelation should be encouraged during training in order to improve the robustness of the model. We also believe that the methodology applied in this work will be beneficial for other researchers investigating Causal NNs and disentangled representations, as the measurements used are flexible in that they permit an extensive range of signal types.

\subsubsection{Limitations}
Our choice of measurements was based on the measurements taken in \citet{content_style_disent_metric}, with the motivation of capturing both signal information content and inter-signal dependency. Nonetheless, other measurement choices are possible. Similarly, the question of exactly which internal model signal to treat as the sampled value of $C$ and $S$ does not have a definite and unique answer for each model and entails some level of qualitative judgement. An exhaustive set of experiments using all possible reasonable choices for these values was infeasible, we have therefore chosen the values which we believe in each case have the closest correspondence to the causal variables employed in each model's design to represent causal and confounder signals. Nonetheless, other researchers might have chosen differently.

It is hard to draw definite conclusions with regard to the results of our analysis with a total of four model architectures trained on three relatively simple datasets. Although promising, more datasets and models should be investigated. 

\subsubsection{Future Work}
An obvious direction of future work is to expand this comparative analysis to include a larger selection of models, tasks, and datasets.

This paper is concerned with measuring the potential benefits of disentangled causal representations for adversarial robustness. Still, other desirable model properties are also of interest, such as out-of-distribution generalisation, few-shot learning, and sample efficiency. We hope that the general disentanglement quantification system utilised in this work will prove useful to other researchers investigating these related topics.

\bibliographystyle{unsrtnat}
\bibliography{biblio}

\input{appendix}

\end{document}

%% file: appendix.tex
\appendix

\begin{table*}[h]
    \centering
    \caption{Metric values for the tested models on MNIST.}\label{tab:metrics_mnist}
    \begin{tabular}{c|c|c|c|c|c}
        \toprule
        Model & $1 - DC(C,S)$ & $DC(X,C)$ & $DC(X,S)$ & $1 - BoI(X,C)$ & $1 - BoI(X,S)$\\
        \midrule
        CAMA & 0.916 & 0.689 & 0.508 & 0.215 & 0.115 \\
        \hline
        CaaM & 0.016 & 0.692 & 0.689 & 0.505 & 0.529 \\
        \hline
        CausalAdv & 0.715 & 0.681 & 0.209 & 0.069 & 0.002 \\
        \hline
        DICE & 0.743 & 0.692 & 0.296 & 0.495 & 0.060 \\
        \bottomrule
    \end{tabular}

    \bigskip\bigskip

    \centering
    \caption{Metric values for the tested models on CIFAR10.}\label{tab:metrics_cifar10}
    \begin{tabular}{c|c|c|c|c|c}
        \toprule
        Model & $1 - DC(C,S)$ & $DC(X,C)$ & $DC(X,S)$ & $1 - BoI(X,C)$ & $1 - BoI(X,S)$\\
        \midrule
        CAMA & 0.930 & 0.331 & 0.808 & 0.042 & 0.507 \\
        \hline
        CaaM & 0.064 &  0.427 & 0.441 & 0.403 & 0.441 \\
        \hline
        CausalAdv & 0.845 & 0.420 & 0.092 & 0.184 & 0.015 \\
        \hline
        DICE & 0.332 & 0.528 & 0.705 & 0.432 & 0.385 \\
        \bottomrule
    \end{tabular}

    \bigskip\bigskip

    \centering
    \caption{Metric values for the tested models on CIFAR100.}\label{tab:metrics_cifar100}
    \begin{tabular}{c|c|c|c|c|c}
        \toprule
        Model & $1 - DC(C,S)$ & $DC(X,C)$ & $DC(X,S)$ & $1 - BoI(X,C)$ & $1 - BoI(X,S)$\\
        \midrule
        CAMA & 0.905 & 0.307 & 0.816 & 0.212 & 0.481 \\
        \hline
        CaaM & 0.315 & 0.416 & 0.349 & 0.336 & 0.409 \\
        \hline
        CausalAdv & 0.896 & 0.470 & 0.083 & 0.317 & 0.002 \\
        \hline
        DICE & 0.196 & 0.681 & 0.761 & 0.438 & 0.382 \\
        \bottomrule
    \end{tabular}

\end{table*}

\section{All Measurement Results}\label{sec:app_all_results}
For completeness, the full set of measurements taken for all twelve trained models is given in Tables \ref{tab:metrics_mnist}, \ref{tab:metrics_cifar10}, and \ref{tab:metrics_cifar100}. These measurement values are averaged across datasets to produce the summarised data in Table \ref{tab:avg_metrics} of the main paper.

\section{Experimental Details}\label{sec:app_exp_setup}
All experiments were performed using the full training epoch budget, and the model with the highest validation accuracy on the clean dataset was returned for testing in each case. For all datasets, the original training split was randomly partitioned into train and validation splits in the ratio $4:1$. 

\subsection{Models}
This subsection gives the implementation details of the four tested models. All models except CAMA were forked and modified from their respective original GitHub repositories, while CAMA was reimplemented from scratch.

\subsubsection{CAMA}
The model was re-implemented using the design described in Appendix C of \citet{CAMA}, using the MNIST design for the MNIST experiments and the CIFAR-binary setup for both our CIFAR10 and CIFAR100 experiments. As perturbed data a horizontal shift of $0.20$ times the image width was applied on all datasets. No finetuning was used during testing.

\subsubsection{CaaM}
The CaaM architecture used is the CNN-CaaM described in \citet{CaaM} with a ResNet18 backbone. All experiments were run with a total training budget of $120$ epochs, using $n=4$ environment splits and the \texttt{auto-iter} environment type for split renewals.

\subsubsection{CausalAdv}
The CausalAdv models were given a total training budget of $120$ epochs, and PGD10 attacks with a maximum perturbation of $\epsilon = 8/255$ were used to generate the adversarial samples.

\subsubsection{DICE}
The training budget was set to $110$ epochs, $200$ epochs, and $250$ epochs for MNIST, CIFAR10, and CIFAR100, respectively. All experiments used PGD10 attacks with a maximum perturbation radius of $\epsilon = 8/255$ for the adversarially perturbed data. The size of the confounder set used for backdoor adjustment was set to 20 for all runs.

\subsection{Attacks}
All attacks were implemented using the \texttt{torchattacks} python library \cite{kim2020torchattacks}. The particular settings for each of the seven attack setups used are described here.

\subsubsection{PGD} The $l_\infty$-bounded PGD attacks used a stepsize of $\alpha=\frac{2}{255}$ for PGD20 and $\alpha=\frac{4}{255}$ PGD40. The $l_2$-bounded PGD attacks used a stepsize of $\alpha=0.2$ for both PGD20 and PGD40. All PGD attacks used a random $\delta$ initialisation.

\subsubsection{CW} Both CW attacks used a $c$-value of $1.0$ in the approximate combined loss function, a confidence value $\kappa=0$, and were optimised with an Adam optimiser with a learning rate of $0.01$.

\subsubsection{FGSM} The only tweakable parameter for the FGSM attack is the maximum perturbation size $\epsilon$ which was set to $\frac{8}{255}$.